\documentclass{article}

\usepackage{arxiv}

\usepackage[utf8]{inputenc} % allow utf-8 input
\usepackage[T1]{fontenc}    % use 8-bit T1 fonts
\usepackage[hidelinks]{hyperref}       % hyperlinks
\usepackage{url}            % simple URL typesetting

\usepackage{amsmath,amssymb,amsfonts,amsthm}

\usepackage{authblk}
\usepackage{booktabs}       % professional-quality tables

\usepackage{amsfonts}       % blackboard math symbols
\usepackage{nicefrac}       % compact symbols for 1/2, etc.
\usepackage{lipsum}		% Can be removed after putting your text content
\usepackage{graphicx}
\usepackage{doi}
\usepackage{float}
\usepackage{stfloats}
\usepackage{subcaption}
\usepackage{textcomp, gensymb}
\usepackage{enumitem}
\usepackage{multirow}
\usepackage{xcolor}
\usepackage{wrapfig}

\title{Small data deep learning methodology for in-field disease detection: case study of potato crops}

\author{
 David Herrera-Poyatos \textsuperscript{*}, Jacinto Domínguez-Rull \textsuperscript{*}, Rosana Montes \textsuperscript{*},  Inés Hernández \textsuperscript{$\dagger$}, Ignacio Barrio \textsuperscript{$\dagger$}, Carlos Poblete-Echeverria \textsuperscript{$\dagger$}, Javier Tardaguila \textsuperscript{$\dagger$}, Francisco Herrera \textsuperscript{*}, Andrés Herrera-Poyatos \textsuperscript{*}
}

\affil{\textsuperscript{*} Andalusian Institute of Data Science and Computational Intelligence (DaSCI), University of Granada, Spain. \\ Emails: \texttt{divadhp@ugr.es}, \texttt{jdrull@correo.ugr.es}, \texttt{rosana@ugr.es}, \texttt{herrera@decsai.ugr.es}, \texttt{andreshp@ugr.es}}

\affil{\textsuperscript{$\dagger$} Televitis Research Group, University of La Rioja, 26006, Logroño, Spain \\ Emails: \texttt{ines.hernandez@unirioja.es}, \texttt{ignacio.barrio@unirioja.es}, \texttt{carlos-alberto.poblete@unirioja.es}, \texttt{javier.tardaguila@unirioja.es}}

\renewcommand{\shorttitle}{In-field disease detection with deep learning}

\hypersetup{
pdftitle={A template for the arxiv style},
pdfsubject={q-bio.NC, q-bio.QM},
pdfauthor={David S.~Hippocampus, Elias D.~Striatum},
pdfkeywords={First keyword, Second keyword, More},
}

\begin{document}

\maketitle

\begin{abstract}
	Early detection of diseases in crops is essential to prevent harvest losses and improve the quality of the final product. In this context, the combination of machine learning and proximity sensors is emerging as a technique capable of achieving this detection efficiently and effectively. For example, this machine learning approach has been applied to potato crops -- to detect late blight (Phytophthora infestans) -- and grapevine crops -- to detect downy mildew.  However, most of these AI models found in the specialised literature have been developed using leaf-by-leaf images taken in the lab, which does not represent field conditions and limits their applicability.

    In this study, we present the first machine learning model capable of detecting mild symptoms of late blight in potato crops through the analysis of high-resolution RGB images captured directly in the field, overcoming the limitations of other publications in the literature and presenting real-world applicability. Our proposal exploits the availability of high-resolution images via the concept of patching, and is based on deep convolutional neural networks with a focal loss function, which makes the model to focus on the complex patterns that arise in field conditions. Additionally, we present a data augmentation scheme that facilitates the training of these neural networks with few high-resolution images, which allows for development of models under the small data paradigm.

    Our model correctly detects all cases of late blight in the test dataset, demonstrating a high level of accuracy and effectiveness in identifying early symptoms. These promising results reinforce the potential use of machine learning for the early detection of diseases and pests in agriculture, enabling better treatment and reducing their impact on crops.
\end{abstract}

% keywords can be removed

\section{Introduction}

Agricultural production plays a significant role in the global economy and food supply. However, it faces serious challenges such as climate change, along with diseases and pests that can devastate entire crops, directly affecting both quality and quantity of the fruit and vegetables produced \cite{syed2022citrus, thakur2022trends, liu2021plant}. Among these diseases, potato late blight (\textit{phytophthora infestans}) is a fungus that manifests as dark spots on the leaves of potato plants. Under humid conditions, a plant affected by late blight can deteriorate rapidly. Additionally, the tubers of infected plants rot quickly when stored for consumption or replanting, making treatment of potato late blight critical for agricultural producers. It is important to note that climate change has intensified the occurrence and pressure of late blight on crops due to changes in temperature and humidity across many agricultural regions \cite{secretariat2021scientific}. Consequently, the use of pesticides to prevent the appearance of late blight has increased considerably, leading to collateral damage such as contamination of food products, as well as significant CO$_2$ emissions from increased pesticide production and a notable rise in pesticide-related costs for farmers \cite{pesticide2022}. For this reason, the European Union has developed legislation to reduce pesticide use by 50\% by 2030 \cite{eutargets2020}.

In this context, the primary tool to minimise pesticide use is the early detection of diseases and pests, so that pesticide treatments can be applied with varying intensities across different areas of the land according to the potential incidence of disease in those zones \cite{cucak2021opportunities}. Conventional methods of visual inspection by farmers to detect diseases and pests, though prevalent, are fraught with challenges such as the labour intensity involved and the subjectivity and potential for human error. Moreover, due to the small size of symptoms, early stages of an infestation can easily go unnoticed, and not all plants are affected simultaneously. Therefore, there is a need to develop automatic and precise detection methods \cite{paulus1997use}.

In an effort to improve the detection and management of agricultural diseases, non-invasive technologies have been explored for crop protection. Proximal sensing and, in particular, RGB image analysis of crops using machine learning has emerged as a vital tool, providing an economical and straightforward alternative for visual disease identification. However, this tool encounters challenges such as interference from external conditions and the high complexity of the images to be analysed. For this reason, previous scientific studies on potato late blight have focused on applying machine learning techniques to synthetic datasets, where images have been taken in a laboratory setting, leaf by leaf \cite{rashid2021multi, mahum2023novel, oppenheim2019using, van2020field}, limiting their applicability. Another line of research involves the use of drones combined with machine learning, but the results have been poor, as the resolution of drone images is insufficient for detecting early symptoms of late blight due to the distance from the crop \cite{sugiura2016field, franceschini2019feasibility}.

From a computer vision perspective, the main difficulty in using `real'' images taken in the field lies in the high complexity of the dataset: these images feature a large number of leaves of various sizes, along with other elements of the plant and the environment. Additional challenges include the small size of the symptoms in each plant and the lack of precise control over the distance at which images are taken. In fact, the problem of disease detection is an example of small object detection, one of the most complex problems in machine learning \cite{liu2021survey}. Indeed, such complexity also makes image labelling an extremely tedious process and, thus, disease detection models have not yet been trained with large amount of data. Therefore, methodologies that tackle this problem should be able incorporate the small data paradigm.

This study introduces the ISD$^4$L methodology that, through the use of high-resolution RGB images acquired in the field, allows for the detection and localisation of diseases and pests in potato crops. Our proposal stands out for the combination of several state-of-the-art computer vision techniques to address the complexities of images taken under field conditions, representing a significant advancement over the state of the art. Namely, the ISD$^4$L methodology employs the concept of patching to exploit high-resolution, incorporates a novel data augmentation scheme that allows our machine learning algorithms to perform under the small data paradigm, and utilises convolutional neural networks with focal loss that focus on more complicated examples during training. The obtained model correctly classifies all high-resolution images of our data set.

The remainder of this paper is organised as follows. In Section \ref{sec:dataset} we introduce the TelevitisPotatoDiseases data set that we use for our experiments. In \ref{sec:methodology} we delve into the ISD$^4$L methodology as well as the new ideas we have introduced in the area of disease detection. In Section \ref{sec:results}, we present the empirical results of our models. Finally, in Section \ref{sec:conclusion}, we highlight the conclusions of this work and suggest potential future research.

\section{The dataset TelevitisPotatoDiseases}
\label{sec:dataset}

\begin{figure}[H]
	\centering
	\begin{subfigure}{.49\textwidth}
		\centering
		\includegraphics[width=.86\linewidth]{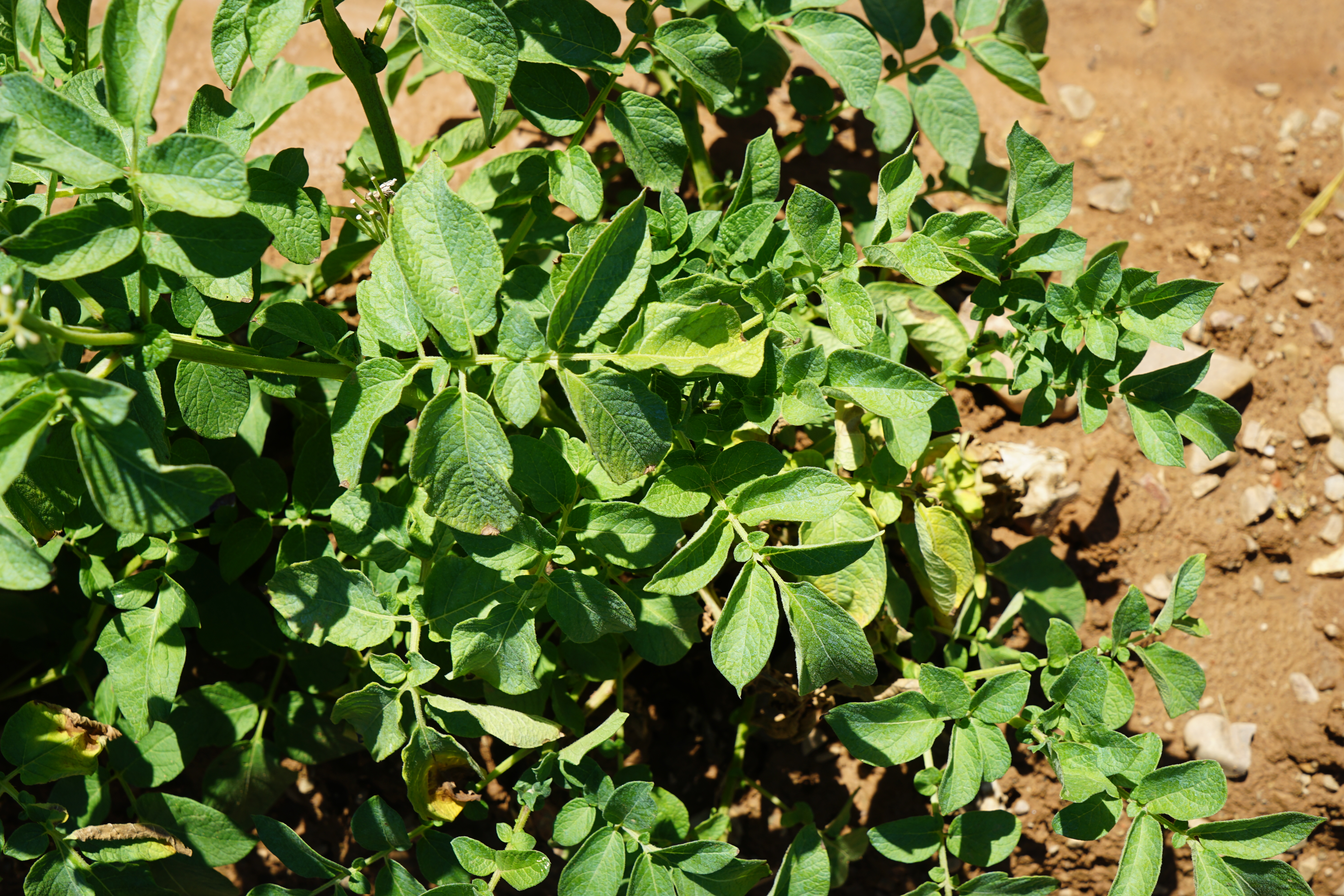}
		\caption{Potato plant with symptoms of late blight}
		\label{sfig:image}
	\end{subfigure}
	\begin{subfigure}{.49\textwidth}
		\centering
		\includegraphics[width=.86\linewidth]{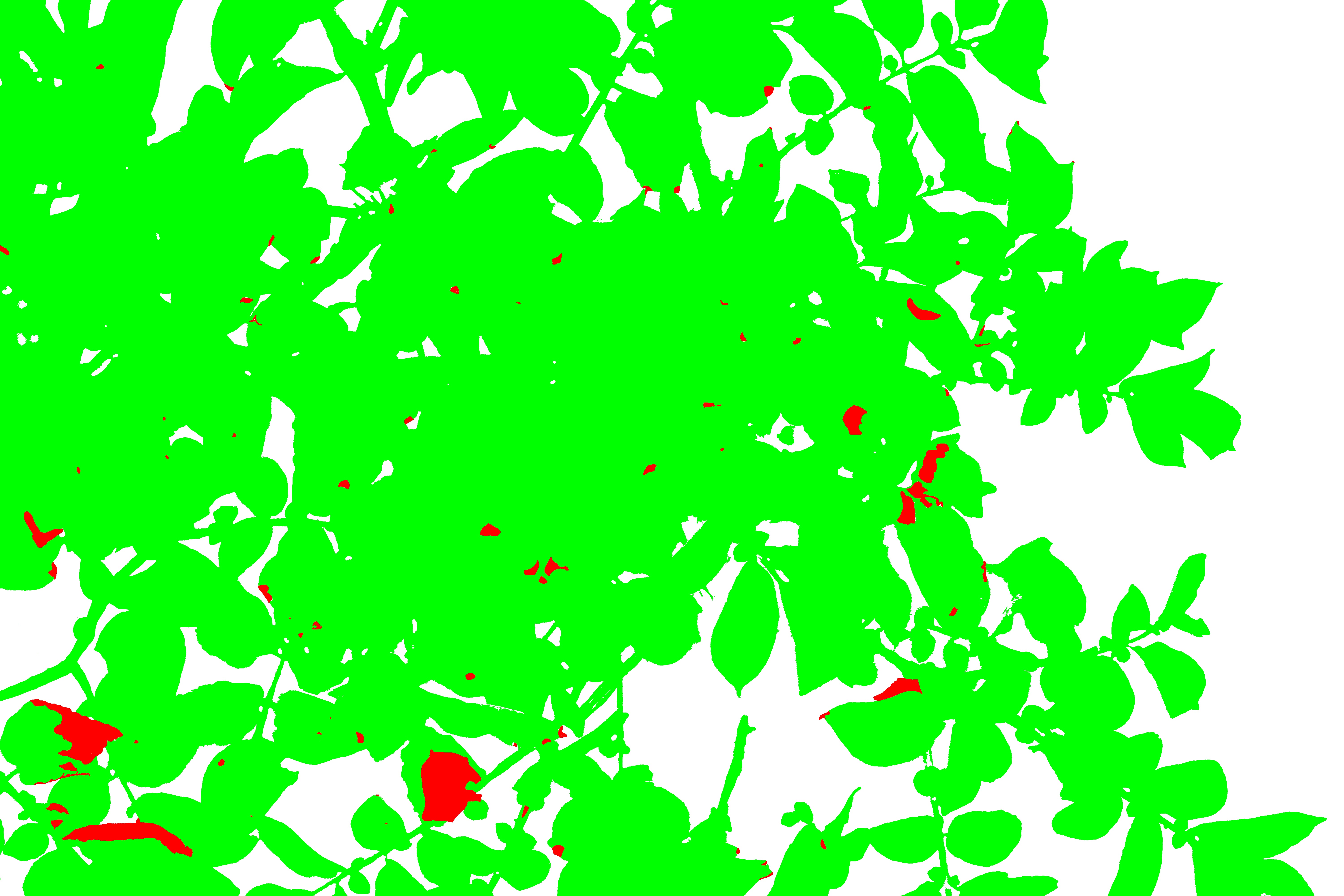}
		\caption{Segmentation of the plant and symptoms}
		\label{sfig:segmentation}
	\end{subfigure}
	\caption{Example of a high-resolution RGB image from our database. The images were taken with the camera pointing towards the ground, from an approximate height of one meter.}
	\label{fig:example-potatoes}
    \end{figure}

In this work, we have created a database of 22 images of potato plants taken in the field, 9 of which show symptoms of late blight (Phytophthora infestans). These images are high-resolution (4000x6000 pixels), providing enough detail to visualise early symptoms of late blight. The images were taken from two experimental commercial potato plots located in northern Spain (Basque Country and La Rioja) during the 2022 season. We used a Sony Alpha 7-II (Sony Corp., Tokyo, Japan) mirrorless RGB digital camera equipped with a Zeiss 24/70 mm lens with optical stabilization. Figure \ref{fig:example-potatoes} shows an image from our dataset. As can be observed, some symptoms are small and difficult to detect with the human eye. For each infected plant, we have created a segmentation map of the plant and symptoms; the information regarding symptoms contained in this map is key for the development of our models.

\section{The ISD$^4$L  methodology: In-field Small Data Disease Detection with Deep Learning}
\label{sec:methodology}

It is well known that traditional CNNs may struggle to detect fine-grained patterns when the original images are massively down-scaled to the network input size. Thus, when high-resolution images are available, splitting these images into patches (smaller subimages) and training models on these patches has become a standard technique in classification based on fine-grained patterns, see, for instance, \cite{herrera2024deep}. This approach has been shown to be promising in the case of disease detection; see \cite{hernandez2023grapevine}. More precisely, in \cite{herrera2024deep} and \cite{hernandez2023grapevine}, high-resolution images in the training set are split into patches to generate the new training set of patches, where further data augmentation may be applied if necessary. Then, at prediction time, new high-resolution images are split into patches, potentially with a sliding window method, and prediction is obtained by aggregating the evaluations of the trained model on these new patches. In the case of disease detection, this aggregation is extremely simple, the high-resolution image presents symptoms of a disease if one of its patches does. We highlight two main issues with this approach:
\begin{enumerate}
    \item \textit{Limited image augmentation.} Once high-resolution images have been split into patches to generate the training set of patches, approaches available in the literature further increase the size of the training set mainly by performing symmetries and rotations on each patch. However, if one performs rotations with an arbitrary angle $\theta$, the rotated patch may  not fit into a square anymore, and thus, the remaining pixels must be filled with blanks, thus losing information and compromising performance of the trained models. Another approach is restricting rotations to the angles $\pi/2, \pi, 3\pi/2$, which limits the amount of data augmentation.
    \item \textit{Amplification of false positives.} A high accuracy of models trained on patches may not translate to a high accuracy of predictions for the high-resolution images. Indeed, if a high-resolution image does not present symptoms of a disease, a false positive on a patch translates to a false positive for the full image prediction. Since each high-resolution image is split into multiple patches at the prediction stage, the probability of a false positive occurring is significantly amplified. We will further delve into this issue in this section.
\end{enumerate}

\begin{figure}[H]
	\centering
    \includegraphics[width=\linewidth]{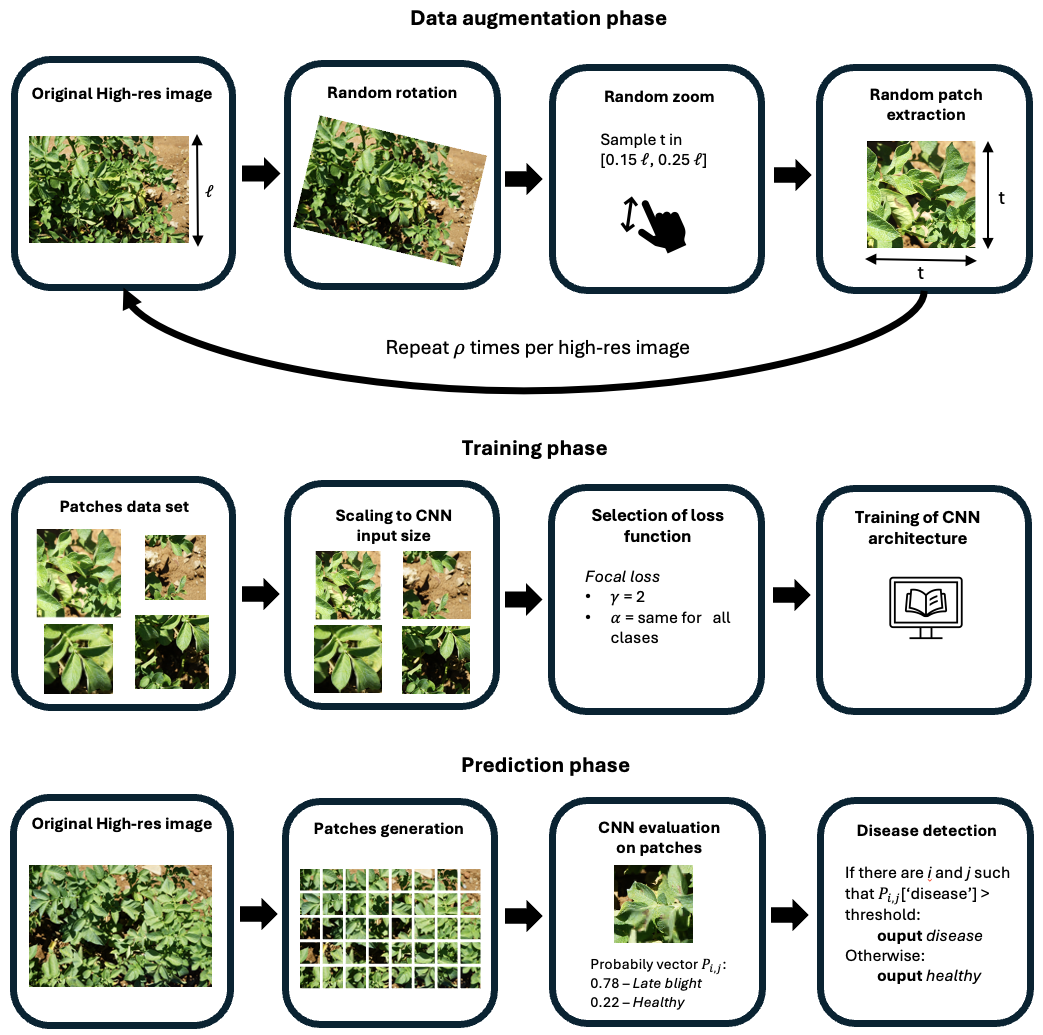}
	\caption{The ISD$^4$L methodology.}
	\label{fig:methodology}
\end{figure}

Given the high cost of labelling and manually segmenting each high-resolution image, one of the main goals of this work is to train a deep learning model that presents high-performance under the small data paradigm. Achieving this would significantly accelerate the adoption of these technologies in practice. Therefore, it is necessary to come up with new approaches that are able to exploit high resolution to generate large training sets of patches. On top of this, the amplification of false positives must be resolved. Our proposal, namely the ISD$^4$L methodology, presents a solution for both problems. There are two key ingredients in our proposal. The first one is a novel data augmentation scheme that is able to extract a high number of patches from each high-resolution image without much redundancy and loss of information. The second one is the application of focal loss as loss function of the CNN architectures which forces the model to focus on the most complicated examples of the training set. Indeed, false positives tend to arise in those patches that present a complex mix of leaves, trunks and back ground. This methodology consists of three main stages, that are illustrated in Figure~\ref{fig:methodology}: the data augmentation stage, the training stage and the prediction stage.

This section is organised as follows. In Section~\ref{sec:methodology:patch-generation} we describe the patch generation algorithm of the TDLI-PIV methodology, in Section~\ref{sec:methodology:training-phase} we describe the prediction phase, delving into the concept of focal loss, and in Section~\ref{sec:methodology:prediction-phase} we describe the prediction phase. %Finally, in Section~\ref{sec:methodology:benefits} we discuss the benefits of our proposal.

\newpage

\subsection{Data augmentation phase: random patch generation} \label{sec:methodology:patch-generation}

As mentioned at the beginning of this section, the main idea to leverage the high resolution of the images in classification based on fine-grained patterns is to extract smaller subimages, on which we train our CNN. These subimages are called \textit{patches}. Here, instead of splitting images into patches and performing data augmentation afterwards, we extract random patches from high-resolution images with different characteristics: size, rotation angle and position. This allows us to maximise the amount of information extracted from each image when generating the training set of patches. Concretely, for each image in the training set, we extract $\rho$ subimages. The parameter $\rho$ can be optimised, and it takes the value $\rho = 200$ in our experiments. Each patch is extracted through the following random process:

\begin{itemize}
    \item \textbf{Rotation}. We rotate the high-definition image by an angle $\theta$, where $\theta$ is a random number between $-\pi$ and $\pi$ (for clarification, here $\pi$ corresponds to $180$ degrees).
    \item \textbf{Zoom}. We randomly select the size of the patch. To do this, we randomly sample number $t$ between $0.15 l$ and $0.25 l$, where $l$ is the number of pixels on the shortest side of the image. In the TelevitisPotatoDiseases data set, $l = 4000$.
    \item \textbf{Random patch extraction}. We crop the rotated high-resolution image to the largest rectangle of pixels that does not include blanks. Then, we select a random square, with $t$ pixels per side, within the rotated high-resolution image. 
    \end{itemize}

In order to label patches as healthy or affected by a disease, we operate as follows. If the patch is extracted from a high-resolution image that shows no symptoms (it has been labelled as healthy), this will be the label for the patch. If the patch has been extracted from a high-resolution image that presents symptoms, then we perform the same transformations used to extract the patch to the segmentation mask of the image, extracting the same patch for this segmentation mask. If the segmentation map for the patch contains symptoms, then the patch is labelled according to the corresponding disease. Otherwise the patch is labelled as healthy.

The use of rotations and zoom ensures that our model is not affected by the angle at which the image is taken or the distance from the sensor to the plant. Moreover, it increases the variety of the patches generated. Due to the randomness of this process, $\rho$ could be significantly large if needed, although this would increase the training time. Our patch extraction algorithm is visualised in Figure~\ref{fig:methodology}, data augmentation phase.

\subsection{Training phase: introducing focal loss} \label{sec:methodology:training-phase}

After data preprocessing, we train a CNN on the patches extracted using the algorithm described in Section \ref{sec:methodology:patch-generation}. First, patches in the training set must be scaled to the input size for our neural network. Indeed, note that the patches generated may have distinct size due to the random zoom operation. Note that the impact downscaling patches is minor compared to the impact of downscaling the full high-resolution image. Here the input size of our CNN model is $380\times380$, whichwas chosen to maximise the information provided to the neural network while maintaining a moderate size for the network layers to avoid overfitting and reduce computation time.

When it comes to the CNN architecture used in our experiments, we selected the EfficientNetV2 deep learning model \cite{tan2021efficientnetv2}, which has shown excellent results in similar problems \cite{hernandez2023grapevine}. The authors of EfficientNetV2 propose seven different architectures depending on the input image size. In this work, we focus on the EfficientNetV2M model, where the input images are 380x380 pixels, to maximize the information available to the neural network while avoiding overfitting. We initialized the weights of EfficientNetV2B3 using transfer learning, leveraging the weights from the model that performed best on ImageNet. To implement our model, we used TensorFlow and Keras \cite{keras}. In our application of EfficientNetV2M, we modified the last three layers, replacing them with a pooling layer, a dense layer with ReLU activation, and, in the final layer, a single neuron with a sigmoid activation function, which predicts the probability that a patch hows symptoms of late blight.

Let $I$ be an instance in our dataset, $c_I \in {0,1}$ the true class of instance $I$, and $p_I$ the probability our model assigns to $I$ belonging to class 1. The CrossEntropy (CE) loss for instance $I$ and prediction $p$ is given by 
\begin{equation*}
    \operatorname{CE}_I(p)= c_I\log(p_I) - (1-c_I) \log(1-p_I).
\end{equation*} 
Traditionally, the CE loss function used in binary classification due to its desirable mathematical properties, as it is derived from the principle of maximum likelihood in statistics \cite[Chapter 5]{Goodfellow-et-al-2016}. However, it has been empirically shown that this loss function is not optimal for problems with class imbalance \cite{johnson2019survey}, as it tends to ignore minority classes. The same reasoning applies to problems where hard-to-classify examples are the minority; in this case, these difficult examples do not sufficiently affect the loss function, and the model ends up making significant errors on them. This behaviour is exacerbated in our problem, where a single false positive with a large error is enough to classify a high-resolution image without symptoms as a plant affected by late blight. 

To avoid false positives with large errors, we propose using the focal loss function with parameters $\alpha = 0.5$ and $\gamma = 2$ \cite{johnson2019survey}. This loss function is a modification of CE that is used in problems with class imbalance. The parameter $\alpha \in (0,1/2]$ is a weight applied to the majority class, while the minority class is weighted by $1-\alpha$. In this case, we keep it at 0.5 as we do not suffer from class imbalance. More important for us is the parameter $\gamma$, which adjusts the weight assigned to an instance based on the error made, such that examples with lower error contribute even less to the loss function. Mathematically, the focal loss function with $\alpha = 0.5$ is given by 
\begin{equation*}
    \operatorname{Focal}_I(p) = - c_I\log(p_I) (1-p_I)^\gamma - (1-c_I) \log(1-p_I) p_I^\gamma.
\end{equation*}
As we will see in the experimentation, $\operatorname{Focal}$ achieves better results than CE by avoiding false positives in our prediction method as false positives are given smaller prediction probabilities compared to classical CE loss.

\subsection{Prediction phase} \label{sec:methodology:prediction-phase}

In this section, we explain how to apply our deep learning model, trained on patches, to detect late blight in a high-resolution image. One initial idea is to divide the high-resolution image into patches using a grid and evaluate our model on them. This algorithm has the following problem: pixels on the edge of a patch tend to have less relevance in the final prediction. To solve this problem, we use the sliding window method. Suppose the high-resolution image has $n \times m$ pixels (in our case $n = 4000$ and $m = 6000$), and we want to extract patches of size $t \times t$, where $t$ is a divisor of $n$. In the sliding window method, for each pair of positive integers $i \in [0, 2(n/t-1)]$ and $j \in [0, 2(\lfloor m/t \rfloor -1)]$, we extract the subimage of size $t$ whose top-left vertex has coordinates $(i t/2, j t/2)$. In our model, we choose $t = n/5 = 800$. In total, we apply the deep learning model to $(2n/t - 1)(2\lfloor m/t \rfloor - 1) = 117$ patches of the original image.

In the remainder of this section, we explain how the CNN results from the patches are aggregated. Recall that, for each subimage $I$, the CNN returns the predicted probability that the subimage shows symptoms of late blight. Let $\hat{p}_I$ denote the output of the network. The probability that the original image presents symptoms of late blight, denoted by $P$, is at least the probability that a particular subimage shows symptoms. Therefore, if the predictions were correct, we could take $\max_I \hat{p}_I$ as a lower bound for $P$. To avoid the effect of a single false positive from the CNN radically affecting our evaluation of the original image, we use a threshold. In our model, we decide that a high-resolution image presents symptoms of late blight if and only if $\max_I \hat{p}_I \ge 0.8$. This approach reduces the impact of local failures on the final model.

%\subsection{Benefits of the proposal}\label{sec:methodology:benefits}

\section{Experimental framework, results and analysis}
\label{sec:results}
The experiment was conducted using the EfficientNetv2M model, which has proven effective in image classification tasks due to its optimized architecture in terms of size and computational efficiency. The model was configured with an input size of \textbf{380x380} pixels, trained for 100 epochs for each fold with a batch size of 32 on a Nvidia A100-SXM4 GPUT with 40GB of memory.

The original dataset TelevitisPotatoDiseases consists of 22 high-resolution images of potato plants. Recall that the parameter $\rho=200$ indicates the number of patches extracted of each high-resolution image. In order to obtain statistical significant results, we have evaluated our proposal using Leave-One-Out validation. Here we briefly discuss this validation approach in our setting, given the fact that we have to differentiate between high-resolution images and patches. First, $\rho$ patches are extracted from each high-resolution image. Then, for each high-resolution image $I$, we train a model using the patches of the rest of the high-resolution images ($21\rho = 4200$ patches in total), and evaluate the resulting model on the patches of $I$, obtaining both an accuracy of prediction at patch level and a prediction for the high-resolution image (which is \textit{Late Blight} if one of the patches has this as a prediction with probability at least $0.8$, and \textit{healthy} otherwise). This iteration is called a \textit{fold}. Table \ref{tab:loo:patch-accuracy} shows the accuracy of the trained CNN on each one of the folds of the leave-one-out validation. We note that most folds present an extremely high accuracy, except for a couple of folds, where accuracy drops due to the fact that there is a significant difference between the image where the model is evaluated and the images in the training set. We hypothesis that this issue would be resolved with a larger training set that represents wider in-field complexities. 

\begin{table}[H]
\centering
    \begin{tabular}{|c|c|}
    \hline
    \textbf{Fold} & \textbf{Accuracy} \\
    \hline
        1 & 1.0000 \\
        2 & 0.9841 \\
        3 & 0.9127 \\
        4 & 0.9921 \\
        5 & 1.0000 \\
        6 & 0.7698 \\
        7 & 0.9921 \\
        8 & 0.9921 \\
        9 & 0.9762 \\
        10 & 1.0000 \\
        11 & 1.0000 \\
        12 & 1.0000 \\
        13 & 0.9841 \\
        14 & 1.0000 \\
        15 & 0.9921 \\
        16 & 1.0000 \\
        17 & 0.9444 \\
        18 & 1.0000 \\
        19 & 0.9921 \\
        20 & 0.9603 \\
        21 & 0.8730 \\
        22 & 1.0000 \\
        \hline
        \textbf{Mean Accuracy} & \textbf{0.9711} \\
        \hline
    \end{tabular}
    \vspace{2mm}
\caption{Results of Leave-One-Out Cross Validation training with focal loss. The accuracy reflects the precision of the patches within the test image.}
\label{tab:loo:patch-accuracy}
\end{table}

Table \ref{tab:loo:full-image} shows the prediction of the leave-one-out validation for each high-resolution image. Recall that a threshold of $0.8$ has been  applied here, that is, a high-resolution image is classified as \textit{Late Blight} if and only if one of the patches has been classified as \textit{Late Blight} by the CNN with probability at least $0.8$. This criterion minimises the incidence of false positives in the overall prediction. The results indicate a significant number of correctly classified images, demonstrating the model's reliability and effectiveness in distinguishing between healthy and infected plants. Indeed, the model correctly classifies 21 out of the 22 images, only making one mistake classifying a healthy plant as \textit{Late Blight} due to committing a false positive on a patch with probability at least $0.8$. 

The results obtained from the leave-one-out cross-validation yield a $0.9545$ accuracy in classifying high-resolution images of potato plants based on the presence or absence of \textit{late blight} symptoms, see Table \ref{tab:loo:full-image}.

\begin{table}[h!]
\centering
\begin{tabular}{|c|c|c|}
\hline
\textbf{Image Class} & \textbf{Correct Prediction} & \textbf{Incorrect Prediction} \\
\hline
\textit{Late Blight} & 9 & 0 \\
Healthy & 12 & 1 \\
\hline
\end{tabular}
    \vspace{2mm}
\caption{Count of images with and without pest and the correctness of the prediction among all folds of Leave-One-Out Cross Validation training with focal loss.}
\label{tab:loo:full-image}
\end{table}

\section{Conclusion}
\label{sec:conclusion}

In this work, we have developed a deep learning model that enables the detection of potato plants infected with late blight in high-resolution RGB images taken directly in the field. Additionally, we envisioned a data augmentation scheme based on patches that allows us to train our models using only a few high-resolution images in the context of small data. Lastly, we demonstrated the utility of a focal loss function for this type of problem, where predictions on the complete image are achieved by aggregating the predictions of a CNN on sub-images.

Our model correctly detects all cases of late blight in the test dataset, indicating a high level of accuracy and effectiveness in detecting the disease under field conditions, and only makes a false positive error on images of healthy plants. Moreover, the accuracy at patch level is 97.11\%. We plan to further increase the complexity of the dataset by including new diseases as well as examples that may challenge the prediction (e.g., plants with leaf damage due to hail, etc.), as well as extending our analysis to other types of crop, such as grapevine, in order to show statistical significant of our results.

These promising results reinforce the potential use of machine learning for the early detection and localisation of diseases and pests in agriculture, which will enable better treatment and reduction of their impact on crops.

\section*{Acknowledgements}

This research results from the Strategic Project IAFER-Cib (C074/23), as a result of the collaboration agreement signed between the National Institute of Cybersecurity (INCIBE) and the University of Granada. This initiative is carried out within the framework of the Recovery, Transformation and Resilience Plan funds, financed by the European Union (Next Generation). In\'es Hern\'andez would like to acknowledge the research funding FPI PhD grant 1150/2020 by Universidad de La Rioja and Gobierno de La Rioja.

\bibliographystyle{unsrt}
\bibliography{references}  %%% Uncomment this line and comment out the ``thebibliography'' section below to use the external .bib file (using bibtex) .

\begin{thebibliography}{10}

\bibitem{syed2022citrus}
Sharifah~Farhana Syed-Ab-Rahman, Mohammad~Hesam Hesamian, and Mukesh Prasad.
\newblock Citrus disease detection and classification using end-to-end anchor-based deep learning model.
\newblock {\em Applied Intelligence}, 52(1):927--938, 2022.

\bibitem{thakur2022trends}
Poornima~Singh Thakur, Pritee Khanna, Tanuja Sheorey, and Aparajita Ojha.
\newblock Trends in vision-based machine learning techniques for plant disease identification: A systematic review.
\newblock {\em Expert Systems with Applications}, page 118117, 2022.

\bibitem{liu2021plant}
Jun Liu and Xuewei Wang.
\newblock Plant diseases and pests detection based on deep learning: a review.
\newblock {\em Plant Methods}, 17:1--18, 2021.

\bibitem{secretariat2021scientific}
M.L. Gullino, R.~Albajes, I.~Al-Jboory, F.~Angelotti, S.~Chakraborty, K.A. Garrett, B.P. Hurley, P.~Juroszek, K.~Makkouk, et~al.
\newblock {\em Scientific review of the impact of climate change on plant pests}.
\newblock FAO on behalf of the IPPC Secretariat, 2021.

\bibitem{pesticide2022}
Heinrich-Böll-Stiftung European~Union et. al.
\newblock Pesticide atlas: Facts and figures about toxic chemicals in agriculture, 2022.
\newblock \url{https://eu.boell.org/en/PesticideAtlas}.

\bibitem{eutargets2020}
European Commission~Food Safety.
\newblock Farm to fork targets, 2020.
\newblock \url{https://food.ec.europa.eu/plants/pesticides/sustainable-use-pesticides/farm-fork-targets-progress_en}.

\bibitem{cucak2021opportunities}
Mladen Cucak, Rafael de~Andrade~Moral, Rowan Fealy, Keith Lambkin, and Steven Kildea.
\newblock Opportunities for improved potato late blight management in the republic of ireland: Field evaluation of the modified irish rules crop disease risk prediction model.
\newblock {\em Phytopathology{\textregistered}}, 111(8):1349--1360, 2021.

\bibitem{paulus1997use}
I~Paulus, R~De~Busscher, and Eddie Schrevens.
\newblock Use of image analysis to investigate human quality classification of apples.
\newblock {\em Journal of Agricultural Engineering Research}, 68(4):341--353, 1997.

\bibitem{rashid2021multi}
Javed Rashid, Imran Khan, Ghulam Ali, Sultan~H Almotiri, Mohammed~A AlGhamdi, and Khalid Masood.
\newblock Multi-level deep learning model for potato leaf disease recognition.
\newblock {\em Electronics}, 10(17):2064, 2021.

\bibitem{mahum2023novel}
Rabbia Mahum, Haris Munir, Zaib-Un-Nisa Mughal, Muhammad Awais, Falak Sher~Khan, Muhammad Saqlain, Saipunidzam Mahamad, and Iskander Tlili.
\newblock A novel framework for potato leaf disease detection using an efficient deep learning model.
\newblock {\em Human and Ecological Risk Assessment: An International Journal}, 29(2):303--326, 2023.

\bibitem{oppenheim2019using}
Dor Oppenheim, Guy Shani, Orly Erlich, and Leah Tsror.
\newblock Using deep learning for image-based potato tuber disease detection.
\newblock {\em Phytopathology}, 109(6):1083--1087, 2019.

\bibitem{van2020field}
Ruben Van De~Vijver, Koen Mertens, Kurt Heungens, Ben Somers, David Nuyttens, Irene Borra-Serrano, Peter Lootens, Isabel Rold{\'a}n-Ruiz, J{\"u}rgen Vangeyte, and Wouter Saeys.
\newblock In-field detection of alternaria solani in potato crops using hyperspectral imaging.
\newblock {\em Computers and electronics in agriculture}, 168:105106, 2020.

\bibitem{sugiura2016field}
Ryo Sugiura, Shogo Tsuda, Seiji Tamiya, Atsushi Itoh, Kentaro Nishiwaki, Noriyuki Murakami, Yukinori Shibuya, Masayuki Hirafuji, and Stephen Nuske.
\newblock Field phenotyping system for the assessment of potato late blight resistance using rgb imagery from an unmanned aerial vehicle.
\newblock {\em Biosystems engineering}, 148:1--10, 2016.

\bibitem{franceschini2019feasibility}
Marston H{\'e}racles~Domingues Franceschini, Harm Bartholomeus, Dirk~Frederik Van~Apeldoorn, Juha Suomalainen, and Lammert Kooistra.
\newblock Feasibility of unmanned aerial vehicle optical imagery for early detection and severity assessment of late blight in potato.
\newblock {\em Remote Sensing}, 11(3):224, 2019.

\bibitem{liu2021survey}
Yang Liu, Peng Sun, Nickolas Wergeles, and Yi~Shang.
\newblock A survey and performance evaluation of deep learning methods for small object detection.
\newblock {\em Expert Systems with Applications}, 172:114602, 2021.

\bibitem{herrera2024deep}
David Herrera-Poyatos, Andr{\'e}s~Herrera Poyatos, Rosa~Montes Soldado, Paloma De~Palacios, Luis~G Esteban, Alberto~Garc{\'\i}a Iruela, Francisco~Garc{\'\i}a Fern{\'a}ndez, and Francisco Herrera.
\newblock Deep learning methodology for the identification of wood species using high-resolution macroscopic images.
\newblock In {\em 2024 International Joint Conference on Neural Networks (IJCNN)}, pages 1--8. IEEE, 2024.

\bibitem{hernandez2023grapevine}
In\'es Hern\'andez, Salvador Guti\'errez, Rub\'en \'Iñigueza, Carlos Poblete-Echeverria, and Javier Tardaguila.
\newblock In-field crop disease symptom localisation using explainable deep learning: use case for downy mildew in grapevine.
\newblock {\em Submitted to Engineering Applications of Artificial Intelligence}, 2023.

\bibitem{tan2021efficientnetv2}
Mingxing Tan and Quoc Le.
\newblock Efficientnetv2: Smaller models and faster training.
\newblock pages 10096--10106, 2021.

\bibitem{keras}
F.~Chollet.
\newblock Keras, 2015.
\newblock \url{https://keras.io}.

\bibitem{Goodfellow-et-al-2016}
Ian Goodfellow, Yoshua Bengio, and Aaron Courville.
\newblock {\em Deep Learning}.
\newblock MIT Press, 2016.

\bibitem{johnson2019survey}
Justin~M Johnson and Taghi~M Khoshgoftaar.
\newblock Survey on deep learning with class imbalance.
\newblock {\em Journal of Big Data}, 6(1):1--54, 2019.

\end{thebibliography}

%%% Uncomment this section and comment out the \bibliography{references} line above to use inline references.
% \begin{thebibliography}{1}

% 	\bibitem{kour2014real}
% 	George Kour and Raid Saabne.
% 	\newblock Real-time segmentation of on-line handwritten arabic script.
% 	\newblock In {\em Frontiers in Handwriting Recognition (ICFHR), 2014 14th
% 			International Conference on}, pages 417--422. IEEE, 2014.

% 	\bibitem{kour2014fast}
% 	George Kour and Raid Saabne.
% 	\newblock Fast classification of handwritten on-line arabic characters.
% 	\newblock In {\em Soft Computing and Pattern Recognition (SoCPaR), 2014 6th
% 			International Conference of}, pages 312--318. IEEE, 2014.

% 	\bibitem{hadash2018estimate}
% 	Guy Hadash, Einat Kermany, Boaz Carmeli, Ofer Lavi, George Kour, and Alon
% 	Jacovi.
% 	\newblock Estimate and replace: A novel approach to integrating deep neural
% 	networks with existing applications.
% 	\newblock {\em arXiv preprint arXiv:1804.09028}, 2018.

% \end{thebibliography}

\end{document}